\documentclass[letterpaper, 10 pt, conference]{ieeeconf}  

\IEEEoverridecommandlockouts                              
\usepackage{floatrow}
\usepackage{comment}
\usepackage[utf8]{inputenc}
\usepackage[T1]{fontenc}
\usepackage{nicefrac}

\usepackage{amsmath} 
\usepackage{amssymb}  
\usepackage{graphicx} 
\usepackage{grffile}
\usepackage[hypcap=true]{caption}
\usepackage{array}
\usepackage{xcolor}
\usepackage{mathtools}
\usepackage{multirow}
\usepackage{multicol}
\usepackage{commath}
\usepackage{colortbl}
\usepackage{enumerate}
\usepackage{hyperref}
\usepackage{changepage}
\usepackage{yhmath}
\usepackage{booktabs}
\usepackage{balance}

\usepackage[font=scriptsize]{caption}

\usepackage[linesnumbered,ruled]{algorithm2e}

\long\def\eo#1{\textcolor{purple}{\bf \small [EO: #1]}}
\long\def\sl#1{\textcolor{pink}{\bf \small [SL: #1]}}
\long\def\jp#1{\textcolor{cyan}{\bf \small [JP: #1]}}

\long\def\methodname{Counter-Hypothetical Particle Filter}
\long\def\methodabbr{CH-PF}

\title{\LARGE \bf
Counter-Hypothetical Particle Filters for Single Object Pose Tracking \\
}

\author{Elizabeth A. Olson \and 
        Jana Pavlasek \and 
        Jasmine A. Berry \and 
        Odest Chadwicke Jenkins%
\thanks{%
E. A. Olson, J. Pavlasek, J. A. Berry, and O. C. Jenkins are with the Robotics Department,
University of Michigan, Ann Arbor, MI, USA,
{\tt \{lizolson, pavlasek, jasab, ocj\}@umich.edu}.}%
}

\begin{document}
\maketitle
\thispagestyle{empty}
\pagestyle{empty}

\setlength{\textfloatsep}{10pt plus 1.0pt minus 2.0pt}
\setlength{\floatsep}{10pt plus 1.0pt minus 2.0pt}
\setlength{\intextsep}{10pt plus 1.0pt minus 2.0pt}
\setlength{\dbltextfloatsep}{5pt plus 1.0pt minus 2.0pt}


\begin{abstract}

Particle filtering is a common technique for six degrees of freedom (6D) pose estimation 
due to its ability to tractably represent belief over object pose.
However, the particle filter is prone to particle deprivation due to the high-dimensional nature of 6D pose. When particle deprivation occurs, it can cause mode collapse of the underlying belief distribution during importance sampling. 
If the region surrounding the true state suffers from mode collapse, recovering its belief is challenging since the area is no longer represented in the probability mass formed by the particles.
Previous methods mitigate this problem by randomizing and resetting particles in the belief distribution, but determining the frequency of reinvigoration has relied on hand-tuning abstract heuristics. 
In this paper, we estimate the necessary reinvigoration rate at each time step by introducing a \textit{Counter-Hypothetical} likelihood function, which is used alongside the standard likelihood. 
Inspired by the notions of plausibility and implausibility from Evidential Reasoning, the addition of our Counter-Hypothetical likelihood function assigns a level of doubt to each particle. The competing cumulative values of confidence and doubt across the particle set are used to estimate the level of failure within the filter, in order to determine the portion of particles to be reinvigorated. 
We demonstrate the effectiveness of our method on the rigid body object 6D pose tracking task.

\end{abstract}

\section{Introduction}











As robot assistants become tasked with accomplishing complex chores, such as preparing a meal or tidying a room, they must be able to interact with a variety of objects.
Object pose estimation in unstructured scenes remains a challenge due to the ambiguity in perception which arises from occlusion and symmetries.  
Particle-based inference methods have been widely applied to six degree of freedom (6D) object pose estimation and tracking due to their ability to represent high-dimensional spaces with finite sample sets
~\cite{Wuthrich2013, sui2015axiomatic, deng2019pose}. 
These methods model estimation uncertainty and are capable of maintaining multiple possible pose hypotheses, which provides robustness in challenging scenarios such as object occlusion and ambiguous symmetries.
\begin{figure}[t]
    \centering
    \includegraphics[width=1.0\linewidth]{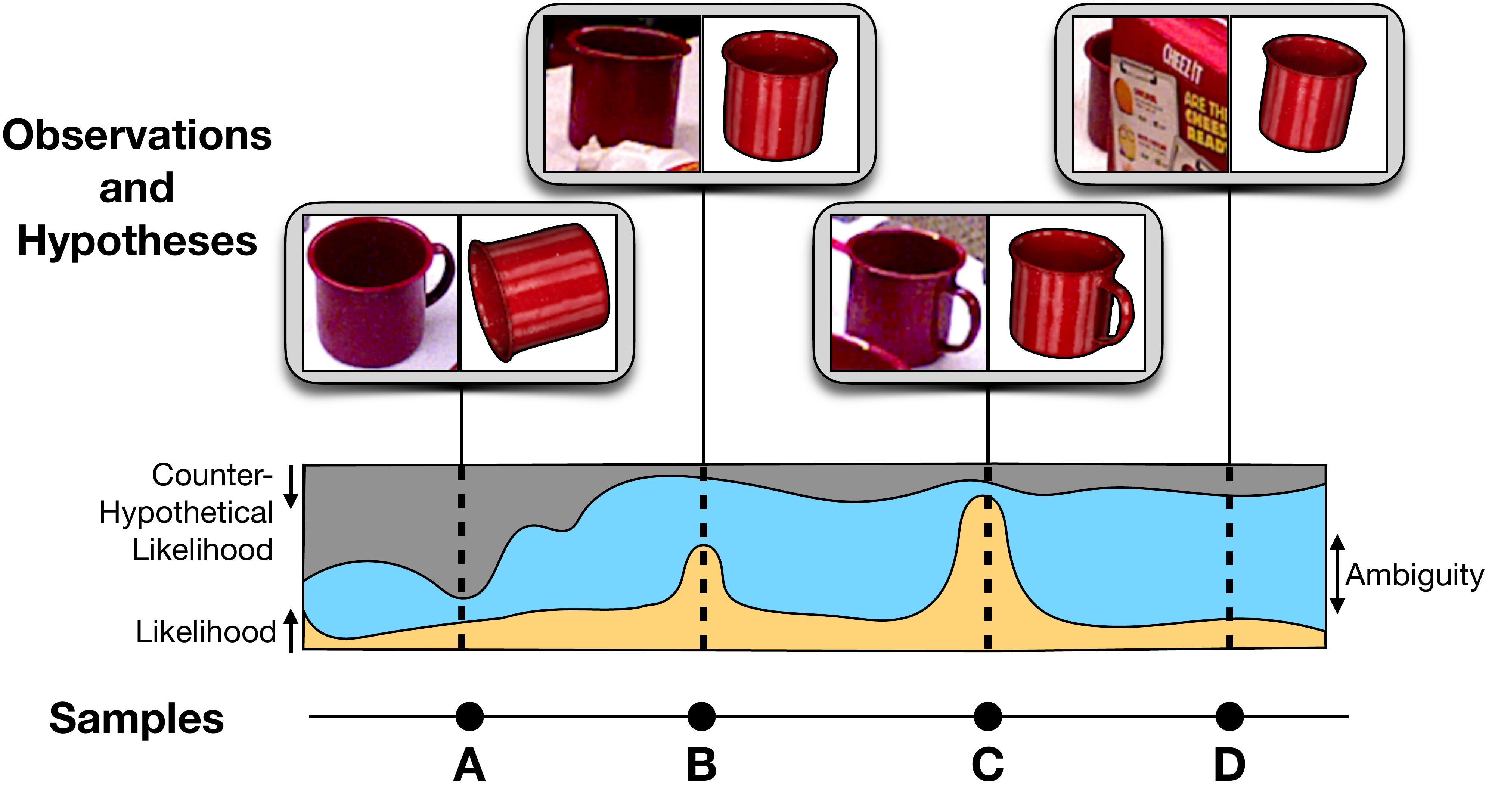}
    \caption{To estimate when the particle filter is failing, we measure the doubt associated with each sample through a Counter-Hypothetical likelihood function. We quantify both the evidence against a given estimate (gray), as well as in support of it (yellow). We estimate these quantities independently of one another, because they are not zero-sum due to ambiguity in the observation (blue). The relative magnitudes of these weightings fluctuate based on the quality of both the observations and estimates, as illustrated by a mug that is (A) unambiguously unlikely (B) plausible yet ambiguous due to the occluded handle (C) unambiguously likely (D) highly ambiguous. }
    \label{fig:overview}
\end{figure}

Despite these promising properties, particle filter algorithms are forced to limit the size of their sample sets to ensure tractability for robotics applications.
When applied to 6D pose tracking, particle filters typically can afford only a small number of samples when compared to the overall size of the continuous state space.
Since the sample set is unable to cover the space completely, certain regions of the state space will contain no particles, making their representation in the belief distribution collapse. This phenomenon is called \textit{particle deprivation}, and can occur due to poor initialization or the stochasticity of importance sampling. Regaining belief in these regions is challenging and can cause the filter to converge to an incorrect local optimum.

One strategy for mitigating particle deprivation is \textit{particle reinvigoration}~\cite{probrob:thrun}, in which reinitialized samples are routinely added to the set. However, determining the portion of particles to be reinvigorated at a given iteration often requires tedious hand-tuning through trial and error. A number of adaptive approaches have been proposed to mitigate this challenge by leveraging the information provided in the likelihood function~\cite{fox1999monte, lenser2000sensor, fox2003adapting}. The likelihood function gives an importance weighting to each sample by measuring the correspondence of the hypothesis to the observed sensor data. However, it only provides a relative weighting of which samples are \textit{better} or \textit{worse}, and no indication of the absolute error in the sample set.

We introduce the Counter-Hypothetical Particle Filter (CH-PF) to counteract the problem of particle deprivation in high-dimensional state spaces. Our method proposes to quantify the confidence that the true state is unrepresented in the sample set.
To this aim, we model the evidence \textit{against} a particular hypothesis, termed the \textit{Counter-Hypothetical} likelihood. Our work measures this weighting independently of the traditional likelihood through the lens of Evidential Reasoning (Dempster-Shafer Theory)~\cite{shafer1976mathematical}. This framework argues that the plausibility and implausiblity of proving an outcome can be based on different factors, and are not zero-sum due to potential overlap and ambiguity of the underlying evidence. Each particle is given both a likelihood and a Counter-Hypothetical likelihood weight. Our method utilizes both likelihoods to quantify the cumulative confidence and doubt across the sample set. The relationship between these values is used to reason about the likelihood that the true state is underrepresented in our sample set, and in turn used to compute an adaptive rate of particle reinvigoration. 

In this paper, we propose the \methodname{}, a particle filtering algorithm designed to mitigate particle deprivation for 6D pose tracking in challenging environments.
We introduce a Counter-Hypothetical likelihood and explain how its utilization alongside the traditional likelihood counteracts particle deprivation by leveraging implausibility information, as illustrated in Figure \ref{fig:overview}. 
We evaluate CF-PF on 6D single object pose tracking on the YCB Video Dataset~\cite{xiang2018posecnn}. Our method achieves better accuracy for cases of high occlusion, particularly when depth data is not available.
\section{Related Work}
\label{sec:related}


\subsection{Object Pose Estimation and Tracking for Robotics}

Pose estimation and tracking have received considerable attention in the robotics community.
In recent years, various works have demonstrated the capability of data-driven methods to provide discriminative pose estimates over a single view~\cite{xiang2018posecnn, tremblay2018deep, wang2019densefusion, he2021ffb6d}, or pose tracking over a sequence of observations~\cite{wen2020se}. 
These methods have achieved impressive results but are prone to inaccuracies, particularly in challenging scenes such as those with heavy clutter.
Probabilistic inference methods instead maintain belief over pose estimates in order to provide additional robustness for robotic manipulation applications~\cite{stoiber2022iterative}. 
We focus on probabilistic inference for object pose estimation and tracking, specifically on particle filtering.

Particle filtering 
is an iterative inference algorithm which can represent an arbitrary nonparametric belief distribution using a set of weighted particles sampled from the state space~\cite{fox1999monte}.
Particle filtering is a common technique for 6D pose estimation and tracking~\cite{sui2015axiomatic, sui2017sum, li2015tracking, choi2012robust} due to its ability to efficiently approximate high-dimensional state spaces and represent multiple competing hypotheses in the belief.
More recently, deep convolutional neural networks (CNNs) have been applied to particle filtering for pose estimation. 
Deng et al.~\cite{deng2019pose} create an observation model from autoencoder embeddings which is used within a Rao-Blackwellized particle filter. Recently, this method has been extended to category-level tracking~\cite{deng2022icaps}. Though the main contribution of our work is the introduction of the Counter-Hypothetical likelihood function, we also demonstrate how it could be similarly learned in an end-to-end fashion. 

To mitigate challenges associated with fully sampling the high-dimensional 6D pose space, previous works have leveraged domain-specific knowledge such as physical constraints~\cite{desingh2016physically}, robotic arm joint angles~\cite{cifuentes2016probabilistic, Wuthrich2013}, or context information~\cite{zeng2020slim}. 
Belief propagation using particle belief representations~\cite{ihler2009particle} has been applied to parts-based object models in order to factor the high-dimensional articulated object localization task~\cite{desingh2019pmpnbp, pavlasek2020parts}.
Similarly, our work addresses the problem of particle deprivation in sampling-based methods for manipulation tasks, but we aim to do so through adaptive reinvigoration, and do not rely on provided object-environment interaction models or parts-based object models.

\subsection{Robust Particle Filtering}



Many works have focused on mitigating particle deprivation. One approach is annealing, in which the distribution of importance weights is smoothed according to a hand-tuned schedule to avoid collapsing modes during importance sampling~\cite{deutscher2000articulated}. 
Pfaff et al. propose an adaptive method to smooth the importance weights using local density estimation around each particle for mobile robot localization~\cite{pfaff2006robust}. 
In contrast, CH-PF does not require modification to the sampling weights, but instead handles particle deprivation through reinvigoration. 

In the global localization stage of Monte Carlo localization for mobile robots, particle deprivation is a common problem~\cite{fox1999monte}, motivating many works to reinitialize samples as needed.  
One approach is to sample from an inverse distribution based on the sensor readings~\cite{thrun2001robust} or ``reset'' a subset of particles when the average likelihood of the sample set is low~\cite{lenser2000sensor}.
Augmented Monte Carlo Localization~\cite{probrob:thrun, gutmann2002experimental} extends this idea by performing particle reinvigoration from a uniform distribution at a rate proportional to the difference between the long- and short-term averages of the particle weights, instead of a fixed threshold.
These methods require a sensor model from which samples can be efficiently drawn, which is challenging to model in RGB images.
Fox et al. proposed modifying the size of the sample set based on the quality of the sample approximation~\cite{fox2003adapting}.
Zhang et al. propose a self-adaptive method that maintains a fixed sample size which is augmented by samples from a ``similar energy region''~\cite{zhang2012self}. This method requires a discretization of the state space. Recent work in localization leverages the estimates of a neural network by sampling from this proposal at a fixed rate and fuses the particles into the distribution through importance sampling~\cite{akai2020hybrid}.

Each of these methods uses the likelihood weights of the particles to estimate the quality of the sample set. \methodabbr{} instead uses a separate source of information, the Counter-Hypothetical likelihood, alongside the likelihood function, to provide an estimate of the overall quality of the particle set. We draw inspiration from Evidential Reasoning~\cite{shafer1976mathematical} for measuring the evidence disproving a hypothetical estimate separately from the likelihood function that looks for supportive evidence. 

\section{Background: Particle Filtering}
\label{sec:pf}

We consider the problem of tracking a known object over time. Given a sequence of RGB images or RGB-D data, $z_{1:t}$, we seek to localize the pose, $x_t \in \mathcal{X}$, of an object at time $t$. We also model any motion to the system, caused by either user input or jittering, with $u_{1:t}$. 
Here $\mathcal{X}$ represents the space of 6D poses, comprised of 3D translation and 3D rotation.

The Bayes filter seeks to model the posterior distribution of the state, $p(x_t\mid x_{1:t-1}, z_{1:t}, u_{1:t})$ by iteratively updating the distribution at each timestep $t$. The posterior is called the \textit{belief} of $x_t$, $bel(x_t)$. 
%
At each time step, the predicted belief, $\widehat{bel}(x_t)$, is obtained by applying the action model to the prior belief distribution. Employing the Markov assumption:
\begin{align}
    \widehat{bel}(x_t) &= \int p(x_t\mid x_{t-1}, u_t) bel(x_{t-1}) dx_t  \label{eq:meas_bayes} 
\end{align}
We can then update this distribution based on the current observation, $z_t$, to estimate the posterior distribution: 
\begin{align}
    bel(x_t) &= p(z_t \mid x_t) \widehat{bel}(x_t) \label{eq:pred_bayes} 
\end{align}

\subsection{Particle Filtering}

The particle filter is a Bayes filtering algorithm in which the belief distribution $bel(x_t)$ 
is a nonparametric distribution approximated by a particle set, $\mathbb{X}_t$:
\begin{align}
    \mathbb{X}_t = \{(x_t^1, \pi_t^1), (x_t^2, \pi_t^2), \ldots , (x_t^N, \pi_t^N)\}
\end{align}
%
Each particle, $x_t^i$ has a corresponding weight, $\pi_t^i$.
The predicted belief in Equation (\ref{eq:meas_bayes}) is formed by applying action $u_t$ to each particle in the previous sample set, $\mathbb{X}_{t-1}$.

The particle set, $\mathbb{X}_{t}$, is generated through \textit{importance sampling}, where the target distribution is $bel(x_t)$ and the proposal distribution is $\widehat{bel}(x_t)$. 
Samples from the proposal are drawn with replacement, in which the probability of a particle being drawn is proportional to its weight, $\pi_t^i$. Typically, the weight is computed using a \textit{likelihood function}, $\mathcal{L}(x_t^i)$, which represents the observation model:
\begin{equation} \label{eq:likelihood}
    \pi_t^i = \mathcal{L}(x_t^i) := p(z_t\mid x_t^i)
\end{equation}
\subsection{Particle Deprivation \& Particle Reinvigoration}
\begin{figure*}[t]
\includegraphics[width=0.95\textwidth]{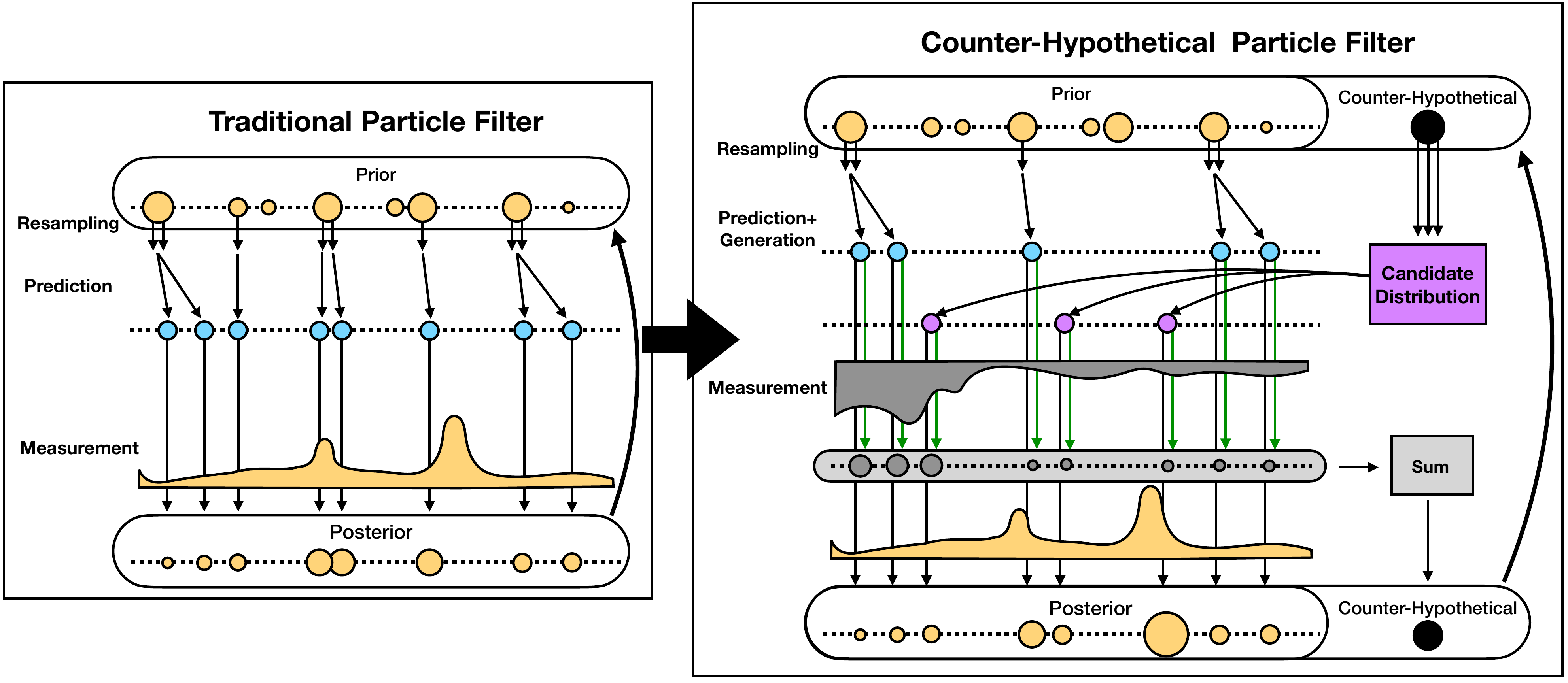}
\caption{An illustration of the proposed modification to the traditional resampling step of the particle filter, visualized as was presented in the Condensation Algorithm work~\cite{isard1998condensation} In the traditional particle filter (left), each iteration begins with a set of weighted particles (top). The samples are drawn with replacement, and then the action model and diffusion are applied to create the prediction distribution (blue). Each sample is then passed through the likelihood function (yellow) to receive a weighting. This posterior distribution then becomes the prior for the next iteration. In our proposed modification (right), each iteration begins with a set of weighted particles (top), as well as a weighting for the Counter-Hypothetical (black). In the resampling stage, only five of the eight particles are created by sampling off the prior distribution (blue), and the Counter-Hypothetical weighting causes three samples to be randomly sampled from the candidate distribution (pink). All of these samples are passed through the Counter-Hypothetical likelihood to be assigned a Counter-Hypothetical weighting (gray). These raw weightings are summed to create a new, singular Counter-Hypothetical weighting representing the doubt across the set. All samples are also passed through the traditional likelihood function (yellow). The posterior distribution and Counter-Hypothetical weighting then become the inputs for the next iteration.}
\label{fig:explanation}
\end{figure*}

If the proposal distribution does not include samples close to the true value of the state, the probability of sampling values in this region is negligibly small. 
This phenomenon is known as \textit{particle deprivation} and is illustrated in Figure~\ref{fig:explanation} (left).
It can occur due to poor initialization, unmodelled movements in the state, or a series of unfortunate draws in importance sampling, causing the particle set to converge to local optima.

A common approach to mitigating particle deprivation is \textit{particle reinvigoration}, in which particles are drawn jointly from the predicted belief, $\widehat{bel}(x_t)$ and a candidate distribution, $\phi_{cand}(x_t)$.
Choices of candidate distribution might include a uniform distribution over the region of interest of the state space, or a wide Gaussian distribution around an initial estimate. This modification allows importance sampling to draw from outside of the sample set, reintroducing samples in underrepresented regions. 
A hyperparameter $\alpha$, where $0 \leq \alpha \leq 1$, controls the proportion of samples to be drawn from $\phi_{cand}(x_t)$.
The final particle set is defined as the union of particles drawn from each set:
%
\begin{align}
    \label{eq:newparticleset}
    \mathbb{X}_t =& 
    \{(x_t^1, \pi_t^1), , \ldots , (x_t^{\alpha N}, \pi_t^{\alpha N})\} 
    \\ &\bigcup  \{(x_t^{\alpha N + 1}, \pi_t^{\alpha N + 1}),  \ldots , (x_t^{N}, \pi_t^{N})\} \nonumber
\end{align}
where $x_t^i \sim  \phi_{cand}$ for $1 \leq i \leq \alpha N$, and $x_t^j \sim \widehat{bel} $ for $\alpha N < j \leq N$. 
Note that in practice, $\alpha N$ is constrained to be an integer.


\section{Counter-Hypothetical Particle Filter}
\label{sec:ch-pf}

Selecting the reinvigoration rate, $\alpha$, is challenging in practice.
Sampling from the candidate distribution too frequently can discard key information from the belief distribution, while sampling from the belief distribution too frequently could lead to particle deprivation. 
\textit{Adaptive particle reinvigoration} mitigates this challenge by determining the frequency with which to draw from each distribution online at each time step.
The Counter-Hypothetical Particle Filter (CH-PF) adaptively selects the reinvigoration rate, $\alpha$, such that it fluctuates in accordance with the portion of the true belief distribution estimated to be underrepresented by $\mathbb{X}_t$.
One method of achieving adaptive particle reinvigoration is Sensor Resetting Localization (SRL)~\cite{lenser2000sensor}. This method defines a probability threshold, $\beta$, which represents a threshold for ``good'' unnormalized likelihood values. The reinvigoration rate is defined as:
\begin{equation}
    \label{eq:srl}
    \alpha = 1 - 
      \left( \frac{1}{\beta N} \sum_{i=1}^N \mathcal{L}(x_t^i) \right)
\end{equation}
CH-PF builds off this equation, using a Counter-Hypothetical likelihood to compute the reinvigoration rate instead of a probability threshold.

\subsection{Counter-Hypothetical Resampling}

To motivate our proposed method for determining the reinvigoration rate, we first rewrite Equation~(\ref{eq:srl}):
\begin{equation}
\label{eq:srl_mod}
\alpha = 1 - 
 \frac{\sum_{i=1}^N \mathcal{L}(x_t^i)}{(\sum_{i=1}^N f(x_t^i)) + (\sum_{i=1}^N \mathcal{L}(x_t^i))}
\end{equation}
where $f(x_t^i):= \beta - \mathcal{L}(x_t^i)$. With this notation, the numerator and right-hand side of the denominator are an aggregate measurement of the likelihood of the sample set. The left-hand side of the denominator, $\sum_{i=1}^N f(x_t^i)$, measures the poor performance across the sample set. 
In this way, calculating the rate of particle reinvigoration in SRL can be seen as simultaneously measuring the positive performance and poor performance of the sample set. However, the measure of poor performance, $f(x_t^i)$, is dependent on the measure of positive performance, as it is defined by $\mathcal{L}(x_t^i)$. This dependency is due to traditional Bayesian probability, in which the observed probability of a state being true and the observed probability of a state being false are always zero-sum. 

\begin{figure}
\centering
\includegraphics[width=0.7\linewidth]{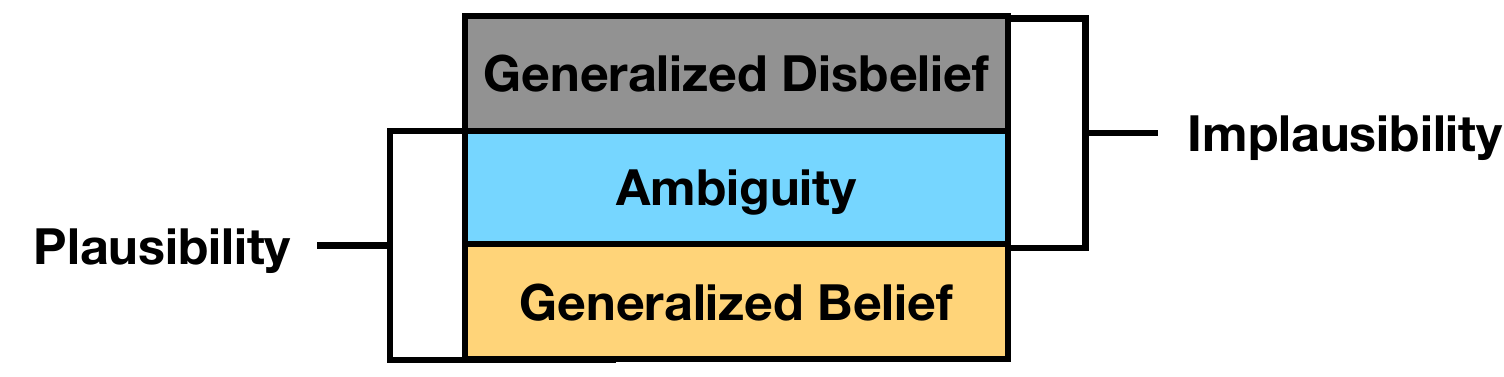}
\caption{Relationship of the quantities in Evidential Reasoning~\cite{shafer1976mathematical}. This paradigm inspires our Counter-Hypothetical likelihood function visualized in Figure~\ref{fig:overview}
}
\label{fig:dst}
\end{figure}

Our method relaxes this assumption by taking inspiration from Evidential Reasoning, also known as Dempster-Shafer Theory~\cite{shafer1976mathematical}. This paradigm allows for the evidence discounting an event, \textit{generalized disbelief}, to be quantified independently of the evidence associated with supporting an event, \textit{generalized belief}. 
As shown in Figure~\ref{fig:dst}, 
Evidential Reasoning models these concepts, as well as the presence of ambiguity or ignorance~\cite{halpern2017reasoning}. 
Generalized belief is a measurement of all evidence that undeniably supports an event and is bounded from above by plausibility, because plausibility includes the ambiguity in the belief. Similarly, generalized disbelief quantifies the evidence that works to disprove an event, while implausiblity is an upper bound that considers ambiguity as well. 


We posit that this framework is apt for our application of evaluating object poses based on images. The occlusions and geometric symmetries present suggest that there is ambiguity in how a given pose is supported or unsupported by the evidence, motivating us to measure these quantities independently. We consider the likelihood function to be analogous to Evidential Reasoning's notion of generalized belief, and therefore introduce a Counter-Hypothetical likelihood to function in accordance with generalized disbelief.%
\footnote{
A common critique of Evidential Reasoning is its appliance to true probability, instead of the probability of provability~\cite{pearl1988probabilistic}. We do not directly use generalized belief to reason about the true underlying probability distribution, we merely take inspiration from Evidential Reasoning by measuring doubt independently of confidence.}



We design the Counter-Hypothetical likelihood to measure how the observed image provides evidence \textit{against} the hypothetical proposed state. Our approach replaces $f(x_t^i)$ from Equation~(\ref{eq:srl}) with a Counter-Hypothetical likelihood, which is estimated independently of $\mathcal{L}(x_t^i)$.
We introduce a function to reason about the confidence of a state counter to our given hypothesis, the \textit{Counter-Hypothetical likelihood}, $\mathcal{C}(x_t)$.
By comparing the quantities of the (unnormalized) likelihoods, $\mathcal{L}(x_t^i)$ and $\mathcal{C}(x_t^i)$, across the proposal distribution, we can reason about the cumulative confidence and doubt in our sample set. 
To this end, we redefine $\alpha$, the reinvigoration ratio, as follows by modifying Equation~(\ref{eq:srl_mod}):
\begin{equation}
     \alpha =  1 - \frac{\sum_{i=1}^N \mathcal{L}(x_t^i)}{ \sum_{i=1}^N \mathcal{C}(x_t^i)+\sum_{i=1}^N  \mathcal{L}(x_t^i)}
\end{equation}
%
We then sample $\alpha N$ particles from $\phi_{cand}$  in accordance with our doubt in the set and sample the remaining $(1-\alpha) N$ particles from $\widehat{bel}$  based on our confidence in the set. 
As such, the Counter-Hypothetical likelihood quantifies our notion of generalized disbelief, which controls the amount of particle reinvigoration to be performed. 

\subsection{Counter-Hypothetical Likelihood for 6D Pose Estimation}

In the context of our application, we design the Counter-Hypothetical likelihood to signal when the pose tracking is in a failure mode. Unlike traditional likelihood functions that must be precisely crafted or trained to ensure the most accurate samples have the highest weights, the Counter-Hypothetical likelihood function can be more crudely or intuitively constructed.  

For a simple example, consider how the captured depth data can be compared against a rendered depth image of a sample at a candidate pose. A traditional likelihood function might assign weightings based on the number of pixels of the object that have rendered and captured depths within a given threshold. This heuristic can be noisy due to the presence of occlusions and difficult to tune. On the other hand, the Counter-Hypothetical likelihood could just measure the number of pixels in which the rendered depth is \textit{less} than the captured depth. The measured depth being significantly closer than the rendered depth can be explained away by a potential occlusion, while the reverse indicates the given
pose is wrong.

In this work, we use deep learning for the Counter-Hypothetical likelihood function and use the encoder architecture of PoseRBPF~\cite{xiang2018posecnn}. We also use their synthetic training data setup. However, instead of training the representation in an auto-encoder manner, we leverage the true pose information present in synthetic data. For our pairs of synthetically rendered scenes and rendered pose candidates, half the training data are positive samples, where the true pose is only slightly perturbed in the candidate rendering. The negative samples are generated by the candidate pose rendering being randomized. These crops are the inputs to duplicates of the encoder network, the output embeddings from which are passed together through three fully connected layers. The network is trained with binary cross-entropy loss.  Through this fashion, the classifier learns to estimate when a hypothetical pose is misaligned with the given observation. At test time, the scores are used as the Counter-Hypothetical weightings. 

\section{Experiments}
\label{sec:exp}

To evaluate the proposed Counter-Hypothetical likelihood, we measure key performance metrics on the YCB-Video Dataset, a benchmarked real-world dataset~\cite{xiang2018posecnn}. We implement a standard particle filter to estimate the 6D pose of a given object across the video sequences. Our results test on both RGB and RGB-D data. 
All of the results are variants of the same particle filter, using the same likelihood function, provided by PoseRBPF~\cite{deng2019pose} and use the same number of particles (50). However each baseline has a different strategy for combating particle deprivation, such as reinvigoration or Rao-Blackwellization. Whenever a candidate distribution is needed for initialization or reinvigoration, a uniform distribution of orientations located in the estimated 2D bounding box from PoseCNN~\cite{xiang2018posecnn} is used. Depth values are sampled from a uniform distribution, but when depth data is present, it is sampled off the measured depth for the object's location. 

\subsection{Baselines}

We compare against methods for particle deprivation specifically designed for 6D pose estimation, as well as adopting other techniques common in mobile robot localization. 

\textbf{Annealing}~\cite{deutscher2000articulated} does not use any particle reinvigoration, but rather has an annealed likelhiood function that cyclically smooths the likelihood weightings.

\textbf{SRL}~\cite{lenser2000sensor} performs adaptive particle reinvigoration from the candidate distribution by comparing the average unnormalized likelihood weighting to a predetermined user threshold. This is a minimum cosine similarity between the embeddings in PoseRBPF.

\textbf{Aug. MCL}~\cite{probrob:thrun, gutmann2002experimental} also performs adaptive particle reinvigoration from the candidate distribution, but the threshold is determined at each time step by user-defined decay rates. 

\textbf{MCL + E2E}~\cite{akai2020hybrid} has a fixed number of samples coming from the predicted distribution with the remaining being sampled off a neural network estimate. We sample off a Gaussian distribution centered at the full 6D pose estimate provided by PoseCNN for the current frame. 

\textbf{PoseRBPF}~\cite{deng2019pose} is run as described in the publication, but with disabling any ground truth information used in the system. The original implementation ensures the initialization is close to the ground truth orientation, or it is completely reset. In our experiment, it is only reset when the likelihood weighting of the estimate drops below a threshold. We also include its suggested variant, \textbf{PoseRBPF++}, in which half the samples are reinvigorated at each time step from the candidate distribution. 

CH-PF, Annealing, SRL, Aug. MCL, and MCL + E2E are filtering across all six dimensions to better test their ability to withstand particle deprivation. Through their Rao-Blackwellized implementation, PoseRBPF and PoseRBPF++ are filtering across only three dimensions of continuous state space (translation), because the orientation space is discretized.

\section{Results}
\label{sec:results}

\begin{figure}
\centering
\includegraphics[width=\linewidth]{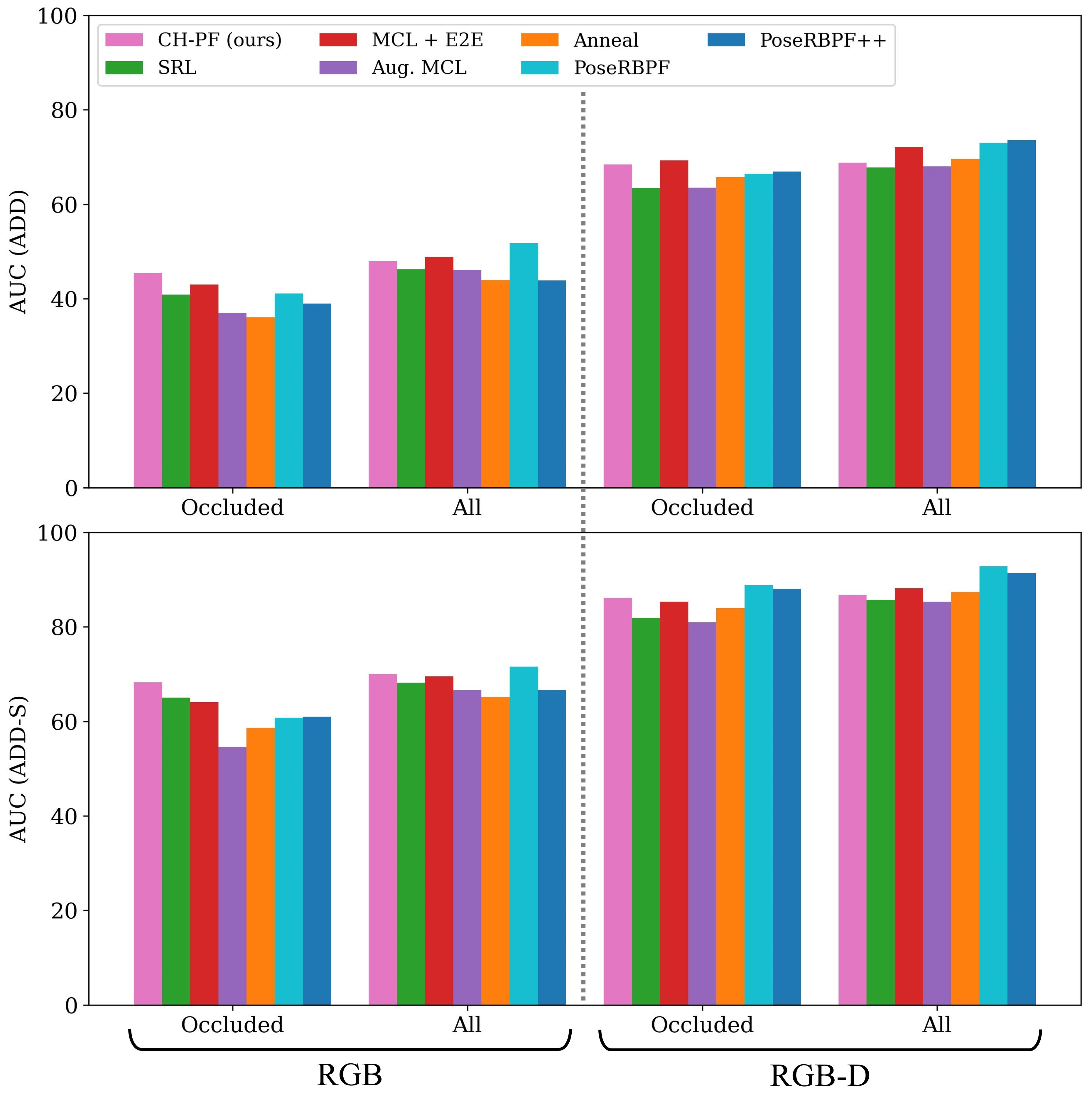}
\caption{\label{fig:results_all} Area Under the Curve scores for all methods for object 6D pose tracking on the YCB Video Dataset. ADD scores and the symmetric version (ADD-S) are presented. Our presented method, CH-PF, outperforms other methods focused on alleviating particle deprivation in instances of occlusion, when the information from the likelihood function is the most noisy (RGB data).}
\end{figure}

We present results with the absolute and symmetric pointwise matching errors between the estimated and ground truth pose (commonly referred to as ADD and ADD-S respectively)~\cite{xiang2018posecnn}. These errors are analyzed by viewing the Area Under the Curve (AUC) score of each method up to 10 cm error, as in other YCB works. Full quantitative results are shown in Figure~\ref{fig:results_all}.  

\begin{figure}
\centering
\includegraphics[width=0.9\linewidth]{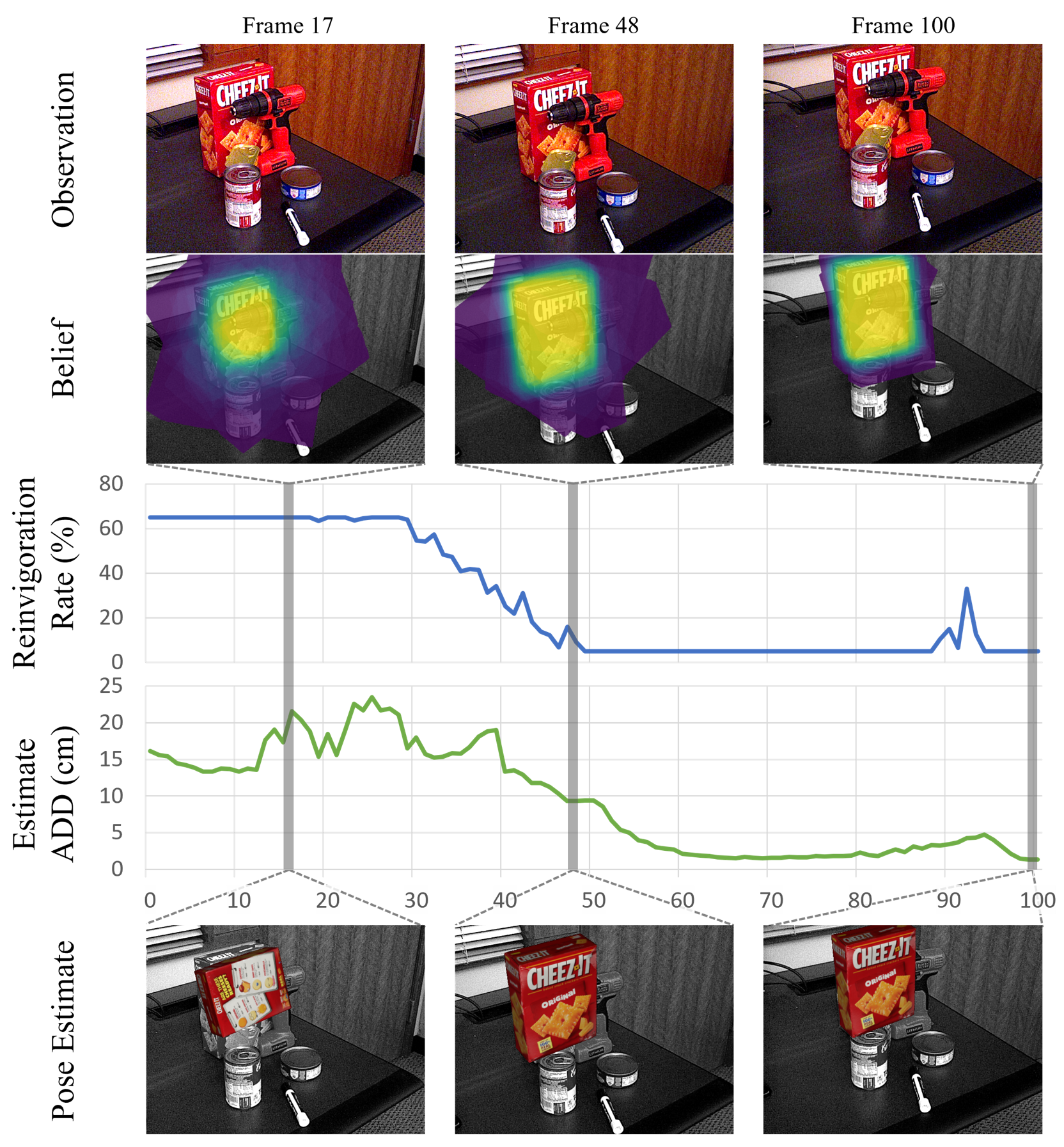}
\caption{\label{fig:qual_cheezits} Selected qualitative results for the \methodname{}. The cracker box is significantly occluded by other objects in the scene. In early iterations (left), the cracker box belief is not converged and the estimate has high error. The reinvigoration rate, calculated from the belief, is high. In later iterations (middle), the belief converges to the ground truth state, and the reinvigoration rate drops. Once the belief has converged (right), the reinvigoration rate is low. This figure is best viewed in color. 
}
\label{fig:qualcracker}

\end{figure}

Our analysis shows that for variants of a particle filter using the same weights but different methods for maintaining particle diversity, there is little deviation in performance across the entire dataset. The Rao-Blackwellized implementations, PoseRBPF and PoseRBPF++, have the best performance.
 However, when looking specifically at the sequences where the given object is occluded, a disparity in accuracy is visible. 
In these cases, PoseRBPF has a decrease in performance. We hypothesize that this is due to the orientation filtering mechanism in PoseRBPF, which makes the system very sensitive to the quality of the initialization; an initial orientation that is flipped is difficult to correct, even with reinvigoration. For this reason, the poor initialization occurring in partially observable scenes lowers its accuracy ranking with respect to the other methods. Aug. MCL can work well when the filter loses track of the pose, but in the case of poor initialization, it does not record a value for an ideal likelihood threshold, and therefore performs little reinvigoration during occlusion. SRL and MCL + E2E  perform on-par with or without occlusions being present. Our method performs similarly to the rest across the entire dataset, but has the highest AUC accuracy for RGB data with occlusions.



\begin{figure}[t!]
\centering
\includegraphics[width=0.9\linewidth]{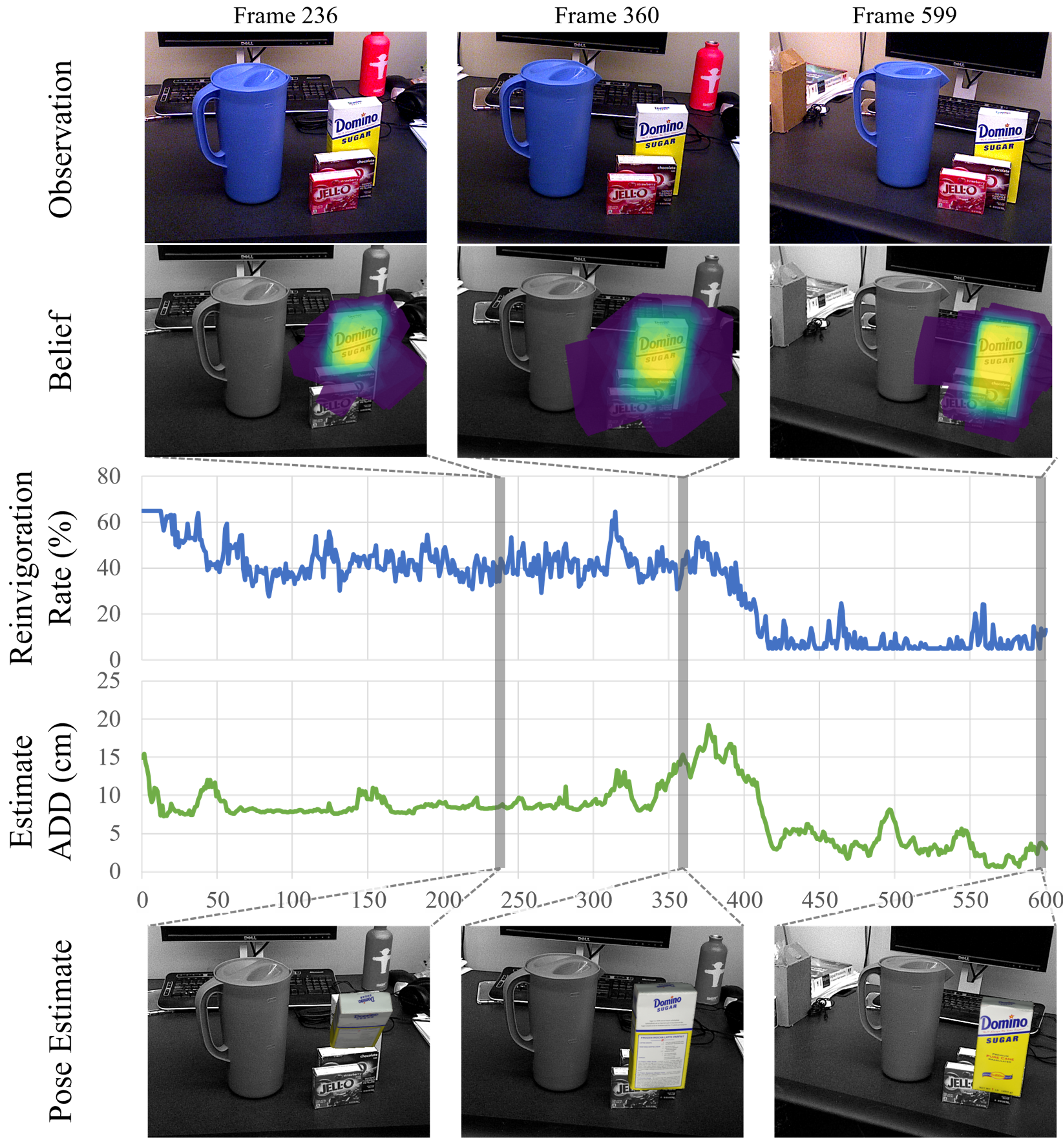}
\caption{\label{fig:qual_sugar} Selected qualitative results for the \methodname{}. The belief of the sugar box converges to a local maximum in early frames (left). \methodabbr{} applies a higher reinvigoration rate to mitigate this. The error in the estimate briefly increases (middle), but the belief eventually converges to the correct estimate (right). 
}
\label{fig:qualsugar}
\end{figure}

Selected qualitative results for the \methodname{} method are shown in Figures \ref{fig:qual_cheezits} and \ref{fig:qual_sugar}. In these examples, the particle filter converged to an incorrect estimate at the beginning of the sequence. While the error was high, the Counter-Hypothetical Particle Filter was able to perform continued global localization with a high percentage of particles performing a coarse search. Once the error dropped, and a plausible region was found, it then reduced the reinvigoration rate to focus its resources on exploring the space more closely.

The main limitation of our work is the additional computation time used at test time to evaluate each sample through an additional likelihood function. For simple heuristics, this would not add much time. In our case of another neural network, it doubles the inference time. Our performance was similar to that of using a single likelihood function for most of the dataset, but improvements in performance for heavily occluded scenes are promising. 


\section{Conclusion}
\label{sec:conclusion}

This work aims to improve the accuracy of particle filters in tracking the 6D pose of rigid objects by adapting the rate of particle reinvigoration based on the estimated incompleteness of the current belief distribution. We propose independently estimating the potential error in each sample through a novel Counter-Hypothetical likelihood function. This modification allows us to reason over the cumulative doubt in our particle set, and use this estimate to apply particle reinvigoration as needed. This paper demonstrates the effectiveness of this modification as it matches overall performance when compared to standard methods of overcoming particle deprivation. Moreover, our particle filter proposed modification improves performance for scenes in heavy occlusion when only RGB data is present.  

Future work will explore more cases for which our proposed incorporation of Evidential Reasoning into Bayesian filters could have more significant increases in performance, such as higher dimensional tracking cases or more dynamic objects and scenes. It will also investigate more computationally effective ways of incorporating this reasoning for real-time inference.

\bibliographystyle{IEEEtran}
\bibliography{references.bib}

\begin{thebibliography}{10}
\providecommand{\url}[1]{#1}
\csname url@samestyle\endcsname
\providecommand{\newblock}{\relax}
\providecommand{\bibinfo}[2]{#2}
\providecommand{\BIBentrySTDinterwordspacing}{\spaceskip=0pt\relax}
\providecommand{\BIBentryALTinterwordstretchfactor}{4}
\providecommand{\BIBentryALTinterwordspacing}{\spaceskip=\fontdimen2\font plus
\BIBentryALTinterwordstretchfactor\fontdimen3\font minus \fontdimen4\font\relax}
\providecommand{\BIBforeignlanguage}[2]{{%
\expandafter\ifx\csname l@#1\endcsname\relax
\typeout{** WARNING: IEEEtran.bst: No hyphenation pattern has been}%
\typeout{** loaded for the language `#1'. Using the pattern for}%
\typeout{** the default language instead.}%
\else
\language=\csname l@#1\endcsname
\fi
#2}}
\providecommand{\BIBdecl}{\relax}
\BIBdecl

\bibitem{Wuthrich2013}
M.~Wuthrich, P.~Pastor, M.~Kalakrishnan, J.~Bohg, and S.~Schaal, ``Probabilistic object tracking using a range camera,'' in \emph{International Conference on Intelligent Robots and Systems (IROS)}.\hskip 1em plus 0.5em minus 0.4em\relax IEEE, 2013, pp. 3195--3202.

\bibitem{sui2015axiomatic}
Z.~Sui, O.~C. Jenkins, and K.~Desingh, ``Axiomatic particle filtering for goal-directed robotic manipulation,'' in \emph{International Conference on Intelligent Robots and Systems (IROS)}.\hskip 1em plus 0.5em minus 0.4em\relax IEEE, 2015, pp. 4429--4436.

\bibitem{deng2019pose}
X.~Deng, A.~Mousavian, Y.~Xiang, F.~Xia, T.~Bretl, and D.~Fox, ``{PoseRBPF}: A {R}ao-{B}lackwellized particle filter for {6D} object pose tracking,'' in \emph{Robotics: Science and Systems (RSS)}, 2019.

\bibitem{probrob:thrun}
S.~Thrun, W.~Burgard, and D.~Fox, \emph{Probabilistic Robotics}.\hskip 1em plus 0.5em minus 0.4em\relax {MIT} Press, 2005.

\bibitem{fox1999monte}
D.~Fox, W.~Burgard, F.~Dellaert, and S.~Thrun, ``Monte carlo localization: Efficient position estimation for mobile robots,'' \emph{AAAI/IAAI}, vol. 1999, no. 343-349, pp. 2--2, 1999.

\bibitem{lenser2000sensor}
S.~Lenser and M.~Veloso, ``Sensor resetting localization for poorly modelled mobile robots,'' in \emph{International Conference on Robotics and Automation ({ICRA})}, vol.~2, 2000, pp. 1225--1232.

\bibitem{fox2003adapting}
D.~Fox, ``Adapting the sample size in particle filters through kld-sampling,'' \emph{The international Journal of robotics research}, vol.~22, no.~12, pp. 985--1003, 2003.

\bibitem{shafer1976mathematical}
G.~Shafer, \emph{A mathematical theory of evidence}.\hskip 1em plus 0.5em minus 0.4em\relax Princeton university press, 1976, vol.~42.

\bibitem{xiang2018posecnn}
Y.~Xiang, T.~Schmidt, V.~Narayanan, and D.~Fox, ``{PoseCNN}: A convolutional neural network for {6D} object pose estimation in cluttered scenes,'' in \emph{Robotics: Science and Systems (RSS)}, 2018.

\bibitem{tremblay2018deep}
J.~Tremblay, T.~To, B.~Sundaralingam, Y.~Xiang, D.~Fox, and S.~Birchfield, ``Deep object pose estimation for semantic robotic grasping of household objects,'' in \emph{Conference on Robot Learning (CoRL)}, 2018.

\bibitem{wang2019densefusion}
C.~Wang, D.~Xu, Y.~Zhu, R.~Mart{\'\i}n-Mart{\'\i}n, C.~Lu, L.~Fei-Fei, and S.~Savarese, ``{DenseFusion}: {6D} object pose estimation by iterative dense fusion,'' in \emph{Conference on Computer Vision and Pattern Recognition ({CVPR})}, 2019, pp. 3343--3352.

\bibitem{he2021ffb6d}
Y.~He, H.~Huang, H.~Fan, Q.~Chen, and J.~Sun, ``{FFB6D}: A full flow bidirectional fusion network for {6D} pose estimation,'' in \emph{Conference on Computer Vision and Pattern Recognition ({CVPR})}, 2021, pp. 3003--3013.

\bibitem{wen2020se}
B.~Wen, C.~Mitash, B.~Ren, and K.~E. Bekris, ``{se(3)-TrackNet}: Data-driven {6D} pose tracking by calibrating image residuals in synthetic domains,'' in \emph{International Conference on Intelligent Robots and Systems (IROS)}.\hskip 1em plus 0.5em minus 0.4em\relax IEEE, 2020, pp. 10\,367--10\,373.

\bibitem{stoiber2022iterative}
M.~Stoiber, M.~Sundermeyer, and R.~Triebel, ``Iterative corresponding geometry: Fusing region and depth for highly efficient {3D} tracking of textureless objects,'' in \emph{Conference on Computer Vision and Pattern Recognition ({CVPR})}.\hskip 1em plus 0.5em minus 0.4em\relax IEEE/CVF, 2022, pp. 6855--6865.

\bibitem{sui2017sum}
Z.~Sui, Z.~Zhou, Z.~Zeng, and O.~C. Jenkins, ``{SUM}: Sequential scene understanding and manipulation,'' in \emph{International Conference on Intelligent Robots and Systems ({IROS})}, 2017, pp. 3281--3288.

\bibitem{li2015tracking}
S.~Li, S.~Koo, and D.~Lee, ``Real-time and model-free object tracking using particle filter with joint color-spatial descriptor,'' in \emph{International Conference on Intelligent Robots and Systems (IROS)}.\hskip 1em plus 0.5em minus 0.4em\relax IEEE, 2015, pp. 6079--6085.

\bibitem{choi2012robust}
C.~Choi and H.~I. Christensen, ``Robust 3d visual tracking using particle filtering on the special euclidean group: A combined approach of keypoint and edge features,'' \emph{The International Journal of Robotics Research}, vol.~31, no.~4, pp. 498--519, 2012.

\bibitem{deng2022icaps}
X.~Deng, J.~Geng, T.~Bretl, Y.~Xiang, and D.~Fox, ``{iCaps}: Iterative category-level object pose and shape estimation,'' \emph{Robotics and Automation Letters (RA-L)}, 2022.

\bibitem{desingh2016physically}
K.~Desingh, O.~C. Jenkins, L.~Reveret, and Z.~Sui, ``Physically plausible scene estimation for manipulation in clutter,'' in \emph{International Conference on Humanoid Robots (Humanoids)}.\hskip 1em plus 0.5em minus 0.4em\relax IEEE, 2016, pp. 1073--1080.

\bibitem{cifuentes2016probabilistic}
C.~G. Cifuentes, J.~Issac, M.~W{\"u}thrich, S.~Schaal, and J.~Bohg, ``Probabilistic articulated real-time tracking for robot manipulation,'' \emph{Robotics and Automation Letters}, vol.~2, no.~2, pp. 577--584, 2016.

\bibitem{zeng2020slim}
Z.~Zeng, A.~R{\"o}fer, and O.~C. Jenkins, ``Semantic linking maps for active visual object search,'' in \emph{International Conference on Robotics and Automation ({ICRA})}.\hskip 1em plus 0.5em minus 0.4em\relax IEEE, 2020.

\bibitem{ihler2009particle}
A.~Ihler and D.~McAllester, ``Particle belief propagation,'' in \emph{Artificial intelligence and statistics}.\hskip 1em plus 0.5em minus 0.4em\relax PMLR, 2009, pp. 256--263.

\bibitem{desingh2019pmpnbp}
K.~Desingh, S.~Lu, A.~Opipari, and O.~C. Jenkins, ``Efficient nonparametric belief propagation for pose estimation and manipulation of articulated objects,'' \emph{Science Robotics}, vol.~4, no.~30, 2019.

\bibitem{pavlasek2020parts}
J.~Pavlasek, S.~Lewis, K.~Desingh, and O.~C. Jenkins, ``Parts-based articulated object localization in clutter using belief propagation,'' in \emph{International Conference on Intelligent Robots and Systems ({IROS})}.\hskip 1em plus 0.5em minus 0.4em\relax IEEE, 2020.

\bibitem{deutscher2000articulated}
J.~Deutscher, A.~Blake, and I.~Reid, ``Articulated body motion capture by annealed particle filtering,'' in \emph{Conference on Computer Vision and Pattern Recognition ({CVPR})}, vol.~2.\hskip 1em plus 0.5em minus 0.4em\relax IEEE, 2000, pp. 126--133.

\bibitem{pfaff2006robust}
P.~Pfaff, W.~Burgard, and D.~Fox, ``Robust {Monte-Carlo} localization using adaptive likelihood models,'' in \emph{European robotics symposium 2006}.\hskip 1em plus 0.5em minus 0.4em\relax Springer, 2006, pp. 181--194.

\bibitem{thrun2001robust}
S.~Thrun, D.~Fox, W.~Burgard, and F.~Dellaert, ``Robust monte carlo localization for mobile robots,'' \emph{Artificial intelligence}, vol. 128, no. 1-2, pp. 99--141, 2001.

\bibitem{gutmann2002experimental}
J.-S. Gutmann and D.~Fox, ``An experimental comparison of localization methods continued,'' in \emph{IEEE/RSJ International Conference on Intelligent Robots and Systems}, vol.~1.\hskip 1em plus 0.5em minus 0.4em\relax IEEE, 2002, pp. 454--459.

\bibitem{zhang2012self}
L.~Zhang, R.~Zapata, and P.~Lepinay, ``Self-adaptive {Monte Carlo} localization for mobile robots using range finders,'' \emph{Robotica}, vol.~30, no.~2, pp. 229--244, 2012.

\bibitem{akai2020hybrid}
N.~Akai, T.~Hirayama, and H.~Murase, ``Hybrid localization using model-and learning-based methods: Fusion of monte carlo and e2e localizations via importance sampling,'' in \emph{International Conference on Robotics and Automation ({ICRA})}.\hskip 1em plus 0.5em minus 0.4em\relax IEEE, 2020, pp. 6469--6475.

\bibitem{isard1998condensation}
M.~Isard and A.~Blake, ``Condensation—conditional density propagation for visual tracking,'' \emph{International journal of computer vision}, vol.~29, no.~1, pp. 5--28, 1998.

\bibitem{halpern2017reasoning}
J.~Y. Halpern, \emph{Reasoning about uncertainty}.\hskip 1em plus 0.5em minus 0.4em\relax MIT press, 2017.

\bibitem{pearl1988probabilistic}
J.~Pearl, \emph{Probabilistic reasoning in intelligent systems: networks of plausible inference}.\hskip 1em plus 0.5em minus 0.4em\relax Morgan kaufmann, 1988.

\end{thebibliography}

\end{document}